\documentclass{elsarticle}

\usepackage{fontenc}
\usepackage{thumbpdf,lmodern}
\usepackage{fancyvrb}
\usepackage{fancyhdr}
\usepackage{framed}
\usepackage{amsmath}
\usepackage{amssymb}
\usepackage{todonotes}
\usepackage{subcaption}
\usepackage{placeins}
\usepackage{booktabs}
\usepackage{graphicx}
\usepackage{color}
\usepackage{ae}
\usepackage{lineno,hyperref}
\usepackage{natbib}
\usepackage{setspace}
\modulolinenumbers[5]

\let\code=\texttt
\newcommand{\pkg}[1]{{\fontseries{b}\selectfont #1}}

\DefineVerbatimEnvironment{Sinput}{Verbatim}{fontshape=sl}
\DefineVerbatimEnvironment{Soutput}{Verbatim}{}
\DefineVerbatimEnvironment{Scode}{Verbatim}{fontshape=sl}

\DefineVerbatimEnvironment{Code}{Verbatim}{}
\DefineVerbatimEnvironment{CodeInput}{Verbatim}{fontshape=sl}
\DefineVerbatimEnvironment{CodeOutput}{Verbatim}{}
\newenvironment{CodeChunk}{}{}
\setkeys{Gin}{width=0.8\textwidth}

\fancypagestyle{plain}{%
   \fancyhf{}
   \fancyfoot[C]{\iffloatpage{}{\thepage}}
   }
\makeatletter
\def\ps@pprintTitle{%
   \let\@oddhead\@empty
   \let\@evenhead\@empty
   \let\@oddfoot\@empty
   \let\@evenfoot\@oddfoot
}
\makeatother
\begin{document}

\begin{frontmatter}
\title{Benchmarking time series classification - Functional data vs machine learning approaches}


\author{Florian Pfisterer}
\author{Xudong Sun\fnref{fn1}}
\author{Laura Beggel\fnref{fn1}}
\author{Fabian Scheipl}
\author{Bernd Bischl}
\address{Department of Statistics, Ludwig-Maximilians-Universit\"at M\"unchen \\ Ludwigstr. 33, \\ 80539, M\"unchen, Germany}

\cortext[1]{Corresponding author: florian.pfisterer@stat.uni-muenchen.de}
\fntext[fn1]{Equal Contribution}

\tnotetext[mytitlenote]{source code available at \url{https://github.com/mlr-org/mlr}}

\begin{abstract}
  Time series classification problems have drawn increasing attention in the machine learning and statistical community. 
  Closely related is the field of functional data analysis (FDA): it refers to the range of problems that deal with the analysis of data that is continuously indexed over some domain. 
  While often employing different methods, both fields strive to answer similar questions, a common example being classification or regression problems with functional covariates. 
  We study methods from functional data analysis, such as functional generalized additive models, as well as functionality to concatenate (functional\-) feature extraction or basis representations with traditional machine learning algorithms like support vector machines or classification trees.
  In order to assess the methods and implementations, we run a benchmark on a wide variety of representative (time series) data sets, with in-depth analysis of empirical results, and strive to provide a reference ranking for which method(s) to use for non-expert practitioners. 
  Additionally we provide a software framework in R for functional data analysis for supervised learning, including machine learning and more linear approaches from statistics. This allows convenient access, and in connection with the machine-learning toolbox \textit{mlr}, those methods can now also be tuned and benchmarked.
\end{abstract}
\begin{keyword}
Functional Data Analysis \sep Time Series \sep Classification
\sep Regression
\end{keyword}
\end{frontmatter}

\newpage
\section{Introduction}
\label{sec:intro}
The analysis of functional data is becoming more and more important in many areas of application such as medicine, economics, or geology (cf. \citet{fdaapplications}, \citet{fda_review}), where this type of data occurs naturally. In industry, functional data are often a by-product of continuous monitoring of production processes, yielding great potential for data mining tasks. A common type of functional data are time series, as time series can often be considered as discretized functions over time.

Many researchers publish software implementations of their algorithms, therefore simplifying the access to already established methods.
Even though such a readily available, broad range of methods to choose from is desirable in general, it also makes it harder for non-expert users to decide which method to apply to a problem at hand and to figure out how to optimize their performance. As a result, there is an increasing demand for automated model selection and  parameter tuning. 

Furthermore, the functionality of available pipeline steps ranges from simple data structures for functional data, to feature extraction methods and packages offering direct modeling procedures for regression and classification. Users are again faced with a multiplicity of software implementations to choose from and, in many instances, combining several implementations may be required. 
This can be difficult and time-consuming, since the various implementations utilize a multiplicity of different workflows which the user needs to become familiar with and synchronize in order to correctly carry out the desired analysis.

There is a wide variety of packages for functional data analysis in R \cite{Rsoft} available that provide functionality for analyzing functional data. Examples range from the \textbf{fda} \citep{fda} package which includes object types for functional data and allows for smoothing and simple regression, to, e.g., boosted additive regression models for functional data in \textbf{FDboost} \citep{fdboost}.
For an extensive overview, see the CRAN task view \citep{crantaskviewfda}. 

Many of those packages are designed to provide algorithmic solutions for one specific problem, and each of them requires the user to become familiar with its user interface. 
Some of the packages, however, such as \pkg{fda.usc} \citep{fda.usc} or \pkg{refund} \citep{refund} are not designed for only one specific analysis task, but combine several approaches. 
Nevertheless, these packages do not offer unified frameworks or consistent user interfaces for their various methods, and most of the packages can still only be applied separately. 


A crucial advantage of providing several algorithms in one package with a unified and principled user interface is that it becomes much easier to compare the provided methods with the intention to find the best solution for a problem at hand.
But to determine the best alternative, one still has to be able to compare the methods at their best performance on the considered data, which requires hyperparameter search and, more preferably, efficient tuning methods.

While the different underlying packages are often difficult and sometimes even impossible to extend to new methods, custom implementations and extensions can be easily included in the accompanying software.



We want to stress that the focus of this paper does not lie in proposing new algorithms for functional data analysis. Its added value lies in a large comparison of algorithms while providing a unified and easily accessible interface for combining statistical methods for functional data with the broad range of functions provided by \pkg{mlr}, most importantly benchmarking and tuning. 
Additionally, the often overlooked possibility of extracting non-functional features from functional data is integrated, which enables the user to apply classical machine learning algorithms such as \textit{support vector machines} \citep{svm} to functional data problems. 

In a benchmark study similar to \citet{bagnall2017great} and \citet{fawaz2019deep}, we explore the performance of implemented methods, and try to answer the following questions:

\begin{enumerate}

    \item Can functional data problems be solved with classical machine learning methods ignoring the functional structure of the data as well as with more elaborate methods designed for this type of data? \citet{bagnall2017great} measure the performance of some non-functional-data-specific algorithms such as the rotation forest \citep{rotationForest}, but this does not yield a complete picture.
    
    \item Guidance on the wide range of available algorithms is often hard to obtain. We aim to make some recommendations in order to simplify the choice of learning algorithm.

    \item Do statistical methods explicitly tailored to the analysis of functional data \citep[e.g. \pkg{FDboost},][]{fdboost} perform well on classical time series tasks? 
    No benchmark results for these methods, which provide interpretable results, are currently available.
    
    \item Many methods that represent functional data in a non-functional domain have been proposed and are also often applied in practice. Examples for this include either hand crafted features \citep[cf.][]{stachl_2015},  summary statistics \citep{tsfeatures}, or generally applicable methods such as wavelet decomposition \citep{wavelets}.
    
    \item Hyperparameter optimization is a very important step in many machine learning applications. In our benchmark, we aim to quantify the impact of hyperparameter optimization for a set of given algorithms on several data sets. 
\end{enumerate}

\paragraph{Contributions}

\noindent As contributions of this paper, we aim to answer the questions posed above.  Additionally, we provide a toolbox for the analysis of functional data. It implements several methods for feature extraction and directly modeling functional data, including a thorough benchmark of those algorithms. This toolbox also allows for full or partial replication of the conducted benchmark comparison.

\newpage 
\section{Related Work}

In the remainder of the paper, we focus on comparing algorithms from the functional data analysis and the machine learning domain. 
Functional data analysis traditionally values interpretable results and valid statistical inference over prediction quality. Therefore functional data algorithms are often not compared  with respect to their predictive performance in literature.
We aim to close this gap. On the other hand, machine learning algorithms often do not yield interpretable results. While we consider both aspects to be important, we want to focus on predictive performance in this paper.



\subsection{Feature extraction and classical machine learning methods}


 
In this work, we differentiate between machine learning algorithms that can directly be applied to functional data, and algorithms intended for scalar features, which we call \textit{classical} machine learning methods.

A popular approach when dealing with functional data is to reduce the problem to a non-functional task by extracting relevant non-functional features \citep{fdaapplications}. 
Applying classical machine learning methods after extracting meaningful features can then lead to competitive results \citep[cf.][e.g.]{goldsmith2014estimator} or at least provide baselines, which are in general not covered by functional data frameworks. In our framework, such functionality is  easily available by combining feature extraction, e.g., based on extracting heuristic properties \citep[cf. \pkg{tsfeatures};][]{tsfeatures} or wavelet coefficients \citep{mallat89, wavelets} and analyzing these derived scalar features with classical machine learning tools provided by \pkg{mlr}.

Based on some existing functionality of the listed packages, we adapt different feature extraction methods. Along with different algorithms already proposed in literature, we propose two new custom methods, \textit{DTWKernel} and \textit{MultiResFeatures}:
\label{sec:dtwkernel}

\begin{itemize}
    \item[] \textbf{tsfeatures}
    \citep{tsfeatures} extracts scalar features, such as auto-correlation functions, entropy and other heuristics from a time series.
    \item[] \textbf{fourier} transforms data from the time domain into the frequency domain using the \textit{fast fourier transform} \citep{fft}.
    Extracted features are either phase or amplitude coefficients.
    
    \item[] \textbf{bsignal} 
    B-Spline representations from package \textit{FDboost} \cite{fdboost} are used as feature extractors. Given the knots vector and effective degree of freedom, we extract the design matrix for the functional data using \textit{mboost}.
    \item[] \textbf{wavelets} \citep{wavelets} applies a discrete wavelet transform to time series or functional data, e.g., with Haar or Daubechies wavelets.
    The extracted features are wavelet coefficients at several resolution levels.
    \item[] \textbf{PCA} projects the data on their principal component vectors. Only a subset of the principal component scores representing a given proportion of signal variance is retained.
    \item[] \textbf{DTWKernel} computes the dynamic time warping distances of functional or time series data to (a set of) reference data.
    We implement \textit{dynamic time warping} (DTW) based feature extraction.
    This method computes the dynamic time warping distance of each observed function to a (user-specified) set of reference curves. The distances of each observation to the reference curves is then extracted as a vector-valued feature. The reference curves can either be supplied by the user, e.g., they could be several typical functions for the respective classes, or they can be obtained from the training data. In order to compute \textit{dynamic time warping distances}, we use a fast dynamic time warping \citep{fastdtw} implementation from package \pkg{rucrdtw} \citep{rucrdtw}.
    \item[] \textbf{MultiResFeatures} extracts features, such as the mean at different levels of resolution (zoom-in steps). Inspired by the image pyramid and wavelet methods, we implement a feature extraction method, \textit{multi-resolution feature extraction} where we extract features like mean and variance computed over specified windows of varying widths. Starting from the full sequence, the sequence is repeatedly divided into smaller pieces, where at each resolution level, a scalar value is extracted. All extracted features are concatenated to form the final feature vector.
\end{itemize}

\subsection{Methods for functional data}

Without feature extraction, direct functional data modeling (both classification and regression) methods incorporated in our package span two families: 
The first family of semi-parametric approaches includes FGAM \citep{refund}, FDboost \citep{fdboost}, and the functional generalized linear model \citep[FGLM;][]{ramsay2006functional}, which are all structured additive models. Those methods use (tensor product) spline basis functions or functional principal components (FPCs) \citep{srivastava2016functional} to represent effects fitted in a generalized additive model. While FGAM and FGLM use the iterated weighted least square (IWLS) method to generate maximum likelihood estimates, FDboost uses component-wise gradient boosting \citep{hothorn2010mboost} to optimize the parameters. Additionally, the estimated parameters can be penalized.  A general formula for this family of methods is $\zeta(Y|X=x)=h(x)=\sum_{j=1}^Jh_j(x)$, where $\zeta$ represents a functional of the conditional response distribution (e.g., an expectation or a quantile), $x$ is a vector of (functional) covariates and $h_j(x)$ are partial additive effects of subsets of $x$ in basis function representation, cf. \citet{greven2017general} for a general introduction. 

The second family of methods are non-parametric methods as introduced in \citet{ferraty2006nonparametric}, e.g., based on (semi-)metrics which quantify local or global differences or distances across curves. For example, the distance between two instances could be defined by the $L_2$ distance of two curves $d(x_i(t), x_j(t)) = \sqrt{\int (x_i(t)-x_j(t))^2dt}$. Kernel functions are used to average over the training instances and weigh their respective contributions based on the value of their distance semi-metric to the predicted instance.
Functional $k$-nearest neighbors algorithms can also be defined based on such semi-metrics.
Implementations can be found in packages \pkg{fda.usc} \citep{fda.usc} and \pkg{classiFunc} \citep{classiFunc}.

\subsection{Toolboxes for functional data analysis}

The package \pkg{fda} \citep{fda} contains several object types for functional data and allows for smoothing and regression for functional data. Analogously, the R-package \pkg{fda.usc} \citep{fda.usc} contains several classification algorithms that can be used with functional data.
In Python, \pkg{scikit-fda} \citep{scikitfda} offers both representation of and (pre-)processing methods for functional data, but only a very small set of machine learning methods for classification or regression problems is implemented at the time of writing. 

As a byproduct of the Time-Series Classification Bake-off \cite{bagnall2017great}, a wide variety of algorithms were implemented and made available. But this implementation emphasizes the benchmark over providing a data analysis toolbox for users, and is therefore not easily usable for inexperienced users.









\subsection{Benchmarks}
The recently published benchmark analysis \textbf{Time-Series Classification Bake-off} by \citet{bagnall2017great} provides an overview of the performance of $18$  state-of-the-art algorithms for time series classification. They re-implement (in Java) and compare 18 algorithms designed especially for time series classification on 85 benchmark time series data sets from \citet{UCRarchive}. In their analysis, they also include results from several standard machine learning algorithms. They note that the \textit{rotation forest} \citep{rotationForest} and \textit{random forest} \citep{breiman01} are competitive with their time series classification baseline \citep[1-nearest neighbor with dynamic time warping distance;][]{somedtw}. Their results show that ensemble methods such as \textit{collection of transformation ensembles} \citep[COTE;][]{COTE} perform best, but for the price of considerable runtime.

Deep learning methods applied to time series classification tasks have also shown competitive prediction power. For example, \cite{fawaz2019deep} provide a comprehensive review of state-of-the-art methods. The authors compared both generative models and discriminative models, including \textit{fully connected neural networks, convolutional neural networks, auto-encoders} and \textit{echo state networks}, whereas only discriminative end-to-end approaches were incorporated in the benchmark study.

The benchmark study conducted in this work does not aim to replicate or compete with earlier studies like  \citep{bagnall2017great}, but instead tries to extend their results.

\section{Functional Data}
\label{sec:fda}

In contrast to non-functional data analysis, where the measurement of a single observation is a vector of scalar components whose entries represent values of the separate multidimensional features, functional data analysis treats and analyses the features themselves as functions over their domain. By learning to represent the underlying function, the carried out analysis is not just restricted to the measured discrete values but it is possible to sample from (and analyze) the entire domain space. 

In this work, we focus on pairs of features and corresponding labels $(x, y)$ for supervised learning.
In contrast to non-functional data analysis, where the measurement of a single observation is a vector of scalar components, functional features are function-valued over their domain. The features $x = (x_1, ..., x_p)$ can thus also be a function, i.e., $x_j = g_j(t)$, 
$g: T \to \mathbb{R}$. 
In practice, functional data comes in the form of observed values $g_j(t), t \in \{1, ..., L\}$, where each $t$ corresponds to a discrete point on the continuum.
Those observed values stem from an underlying function $f$ evaluated over a set of points.
A frequent type of functional data is time series data, i.e., measurements of a process measured at discrete time-points.

\begin{figure}[ht]
\centering
\includegraphics[width=0.9\textwidth]{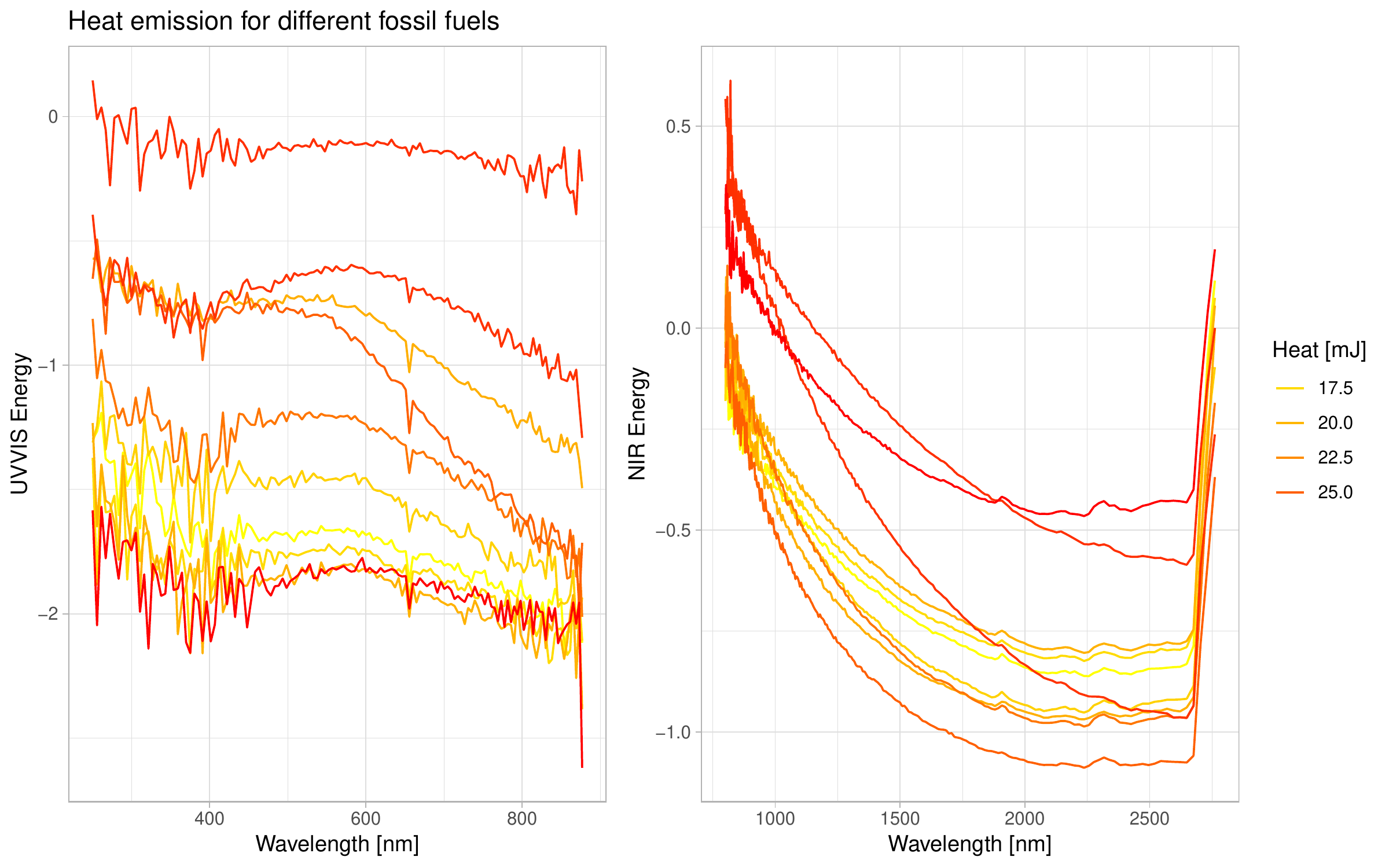}
\caption{Scalar-on-function regression:
         Spectral data for fossil fuels \citep{fueldata}}
\centering
\label{fig:fdboost}
\end{figure}

For example, in some electrical engineering applications, signals are obtained over time at a certain sampling rate, but other domains are possible as well. Spectroscopic data, for example, are functional data recorded over certain parts of the electromagnetic spectrum.
One such example is depicted in Figure \ref{fig:fdboost}. It shows spectroscopy data of fossil fuels \citep{fueldata} where the measured signal represents reflected energies in the ultraviolet-visible (UV-VIS) and the near infrared spectrum (NIR). 
In the plot, different colors correspond to different instances. This is a typical example of a scalar-on-function regression problem, where the inputs are a collection of spectroscopic curves for a fuel, and the prediction target is the heating value of the fossil fuel. 

\begin{figure}[ht]
\begin{subfigure}{.5\textwidth}
  \centering
  \includegraphics[width=.9\linewidth]{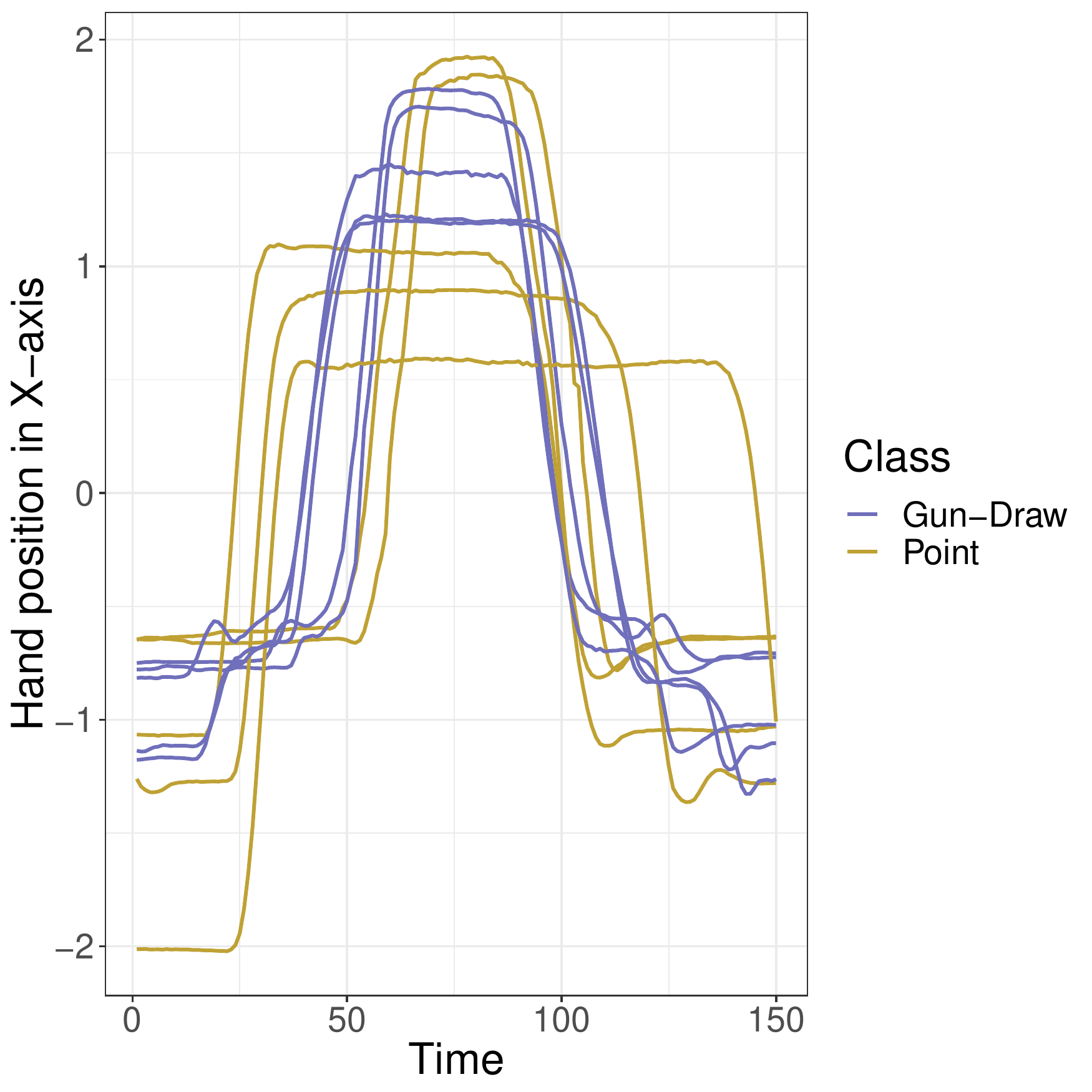}
  \caption{Gunpoint}
  \label{fig:gunpoint}
\end{subfigure}%
\begin{subfigure}{.5\textwidth}
  \centering
  \includegraphics[width=.9\linewidth]{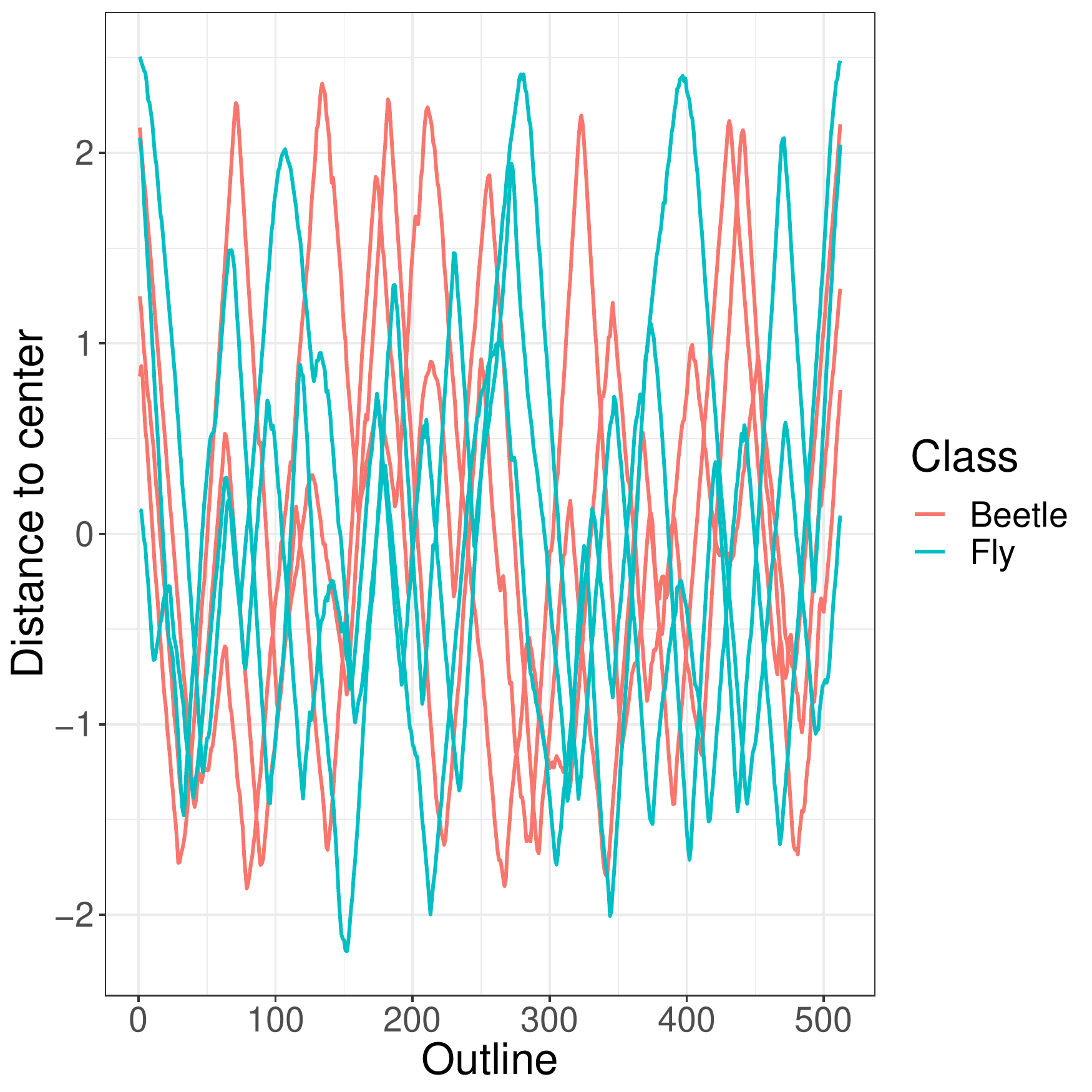}
  \caption{BeetleFly}
  \label{fig:beetlefly}
\end{subfigure}
\caption{Excerpts from two time series classification data sets. (a): Gunpoint data \citep{gunpoint}, (b): BeetleFly Data \citep{beetlefly}.}
\label{fig:tsc}
\end{figure}

In Figure~\ref{fig:tsc}, we display two functional classification scenarios. The goal in those scenarios is to distinguish the class type of the curve, which can also be understood as a function-on-scalar problem. Figure \ref{fig:gunpoint} shows the vertical position of an actor's hand while either drawing a toy-gun and aiming at a target, or just imitating the motion with the blank hand. This position is measured over time. The two different types of classes of the curves can be distinguished by the color scheme. \\
Figure \ref{fig:beetlefly} shows a data set built for distinguishing images of beetles from images of flies based on their outlines. While following the outline, the distance to the center of the object is measured which is then used for classification purposes. 
The latter data sets are available from \cite{UCRarchive}.

The interested reader is referred to \citet{ramsay2006functional} and \citet{kokoszkareimherr2017} for more in-depth introductions to this topic.

\section{Functional Data Analysis with mlrFDA}

Along with the benchmark, we implement the software \pkg{mlrFDA}, which extends the popular machine learning framework \pkg{mlr} for the analysis of functional data.
As the implemented functionality is an extension of the \pkg{mlr} package, all of the functionality available in \pkg{mlr} transfers to the newly added methods for functional data analysis.
We include a brief overview of the implemented functionality in \ref{sec:fda_models}.
A more in-detail overview and tutorial on \pkg{mlr} can be found in the mlr tutorial \citep{mlrtutorial}.

\pkg{mlr} provides a unified framework for machine learning methods in R, currently supporting \emph{tasks} from $4$ main problem types: (multilabel-)classification, regression, cluster analysis, and survival analysis. For each problem type, many algorithms (called \emph{learners}) are integrated. This yields an extensive set of modeling options with a unified, simple interface. Moreover, advanced techniques such as hyperparameter tuning, preprocessing and feature selection are also part of the package. An additional focus lies on extensibility, allowing the user to integrate their own algorithms, performance measures and preprocessing methods.
As \pkg{mlrMBO} seamlessly integrates into the new software, many different tuning procedures can be readily adapted by the user.
Tuning of hyperparameters is usually not integrated in software packages for functional data analysis and thus would require the user to write additional, non-trivial code that handles (nested) resampling, evaluation and optimization methods.

\pkg{mlrFDA} contains several functional data algorithms from several R packages, e.g., \pkg{fda.usc}, \pkg{refund} or \pkg{FDboost}. The algorithms' functionality, however, remains unchanged, only their user interface is standardized for use with \pkg{mlr}. 
For detailed insights into the respective algorithms, full documentation is available in the respective packages.
 
Since our toolbox is built on \pkg{mlr}'s extensible class system, our framework is easily extensible to other methods that have not yet been integrated, and the user can include his or her own methods which do not necessarily need to be available as a packaged implementation. 
Additionally, \pkg{mlrFDA} inherits \pkg{mlr}'s functionality for performance evaluation and benchmarking, along with extensive and advanced (hyperparameter) tuning. This makes our platform very attractive for evaluating which algorithm fits best to a problem at hand, and even allows for large benchmark studies.

\section{Benchmark Experiment}

In order to enable a comparison of the different approaches, an extensive benchmark study is conducted.
This paper does not aim to replicate or reproduce results obtained by \citet{bagnall2017great} or \citet{fawaz2019deep}.
Instead we focus on providing a benchmark complementary to previous benchmarks. This is done because $i)$ the experiments require large amounts of computational resources, and $ii)$ the added value of an exact replication of the experiments (with open source code) is comparatively small.
Nonetheless, we aim for results that can be compared, and thus extend the results obtained by \citet{bagnall2017great} by staying close to their setup. The experiments were carried out on a high performance computing cluster, supported by the Leibniz Rechenzentrum Munich. Individual runs were allowed up to $2.2$ GB of RAM and $4$ hours run-time for each evaluation.
We want to stress that this benchmark compares \textit{implementations}, which does not always necessarily correspond to the performance of the corresponding theoretical \textit{algorithm}. Additionally, methods for functional data analysis are traditionally more focused on valid statistical inference and interpretable results, which does not necessarily coincide with high predictive performance.

\subsection{Benchmark Setup}

\begin{table}[ht]
\begin{small}
\scalebox{0.9}{
\begin{tabular}{|l|l|}
	\hline
	\textbf{Data sets} &  51 Data sets, see table \ref{tab:datasets} \\
	
	\hline
	\textbf{Algorithms} & Function (Package)\\
	                   
	Machine Learning: & - \texttt{glmnet} (\pkg{glmnet})\\
	 & - \texttt{rpart} (\pkg{rpart}) \\
	 & - \texttt{ksvm}$^\star$ (\pkg{kernlab}) \\
	 & - \texttt{ranger}$^\star$ (\pkg{ranger}) \\
	 & - \texttt{xgboost}$^\star$ (\pkg{xgboost}) \\
	 & \\
	Functional Data & - \texttt{classif.knn}(\pkg{fda.usc}) \\
	 & - \texttt{classif.glm} (\pkg{fda.usc})\\
	 & - \texttt{classif.np} (\pkg{fda.usc})\\
	 & - \texttt{classif.kernel}(\pkg{fda.usc}) \\
	 & - \texttt{FDboost} (\pkg{FDboost}) \\
	 & - \texttt{fgam} (\pkg{refund})\\
	 & - \texttt{knn with dtw} (\pkg{classiFunc})\\
	 & \\
	Feature Extraction + ML & - feature extraction: see table \ref{tab:feature_extractors}\\
     & - in combination with ML algorithms marked with a $^\star$.\\
	\hline
	\textbf{Measures} & mean misclassification error, training time \\
	\hline
	\textbf{Resampling} & 20-fold stratified sub-sampling;\\
	& class balances and train/test set size as in \cite{bagnall2017great}.\\
	\hline
	\textbf{Tuning} & 100 iterations of Bayesian optimization (3-fold inner CV).\\
	                & Corresponding hyperparameter-ranges can be obtained\\
	                & from tables \ref{tab:parsets_fda_extractors} and \ref{tab:ps_algo_tuned}. \\
	\hline
\end{tabular}}
\end{small}
\caption{Benchmark experiment setup}
\label{tab:benchmark}
\end{table}

\noindent A benchmark experiment is defined by four important characteristics: The \textit{data sets} algorithms are tested on, the \textit{algorithms} to be evaluated, the \textit{measures} used for evaluating predictive performance, and a \textit{resampling strategy} used for generating train and test splits of the data. 
A comprehensive overview of the conducted benchmark setup can be obtained from Table \ref{tab:benchmark}.

These characteristics are briefly described subsequently before providing and discussing the results.
We use a subset of 51 data sets from the popular UCR archive \citep{UCRarchive} in order to enable a comparison of results in \cite{bagnall2017great} with the additional methods described in this paper.
The data sets stem from various application types such as ECG measurements, sensor data, or image outlines, therefore having varying training set sizes or measurement lengths. 
For more detailed information about the data sets, see \citet{UCRarchive}.

We selected data using the following criteria:
In order to reduce the computational resources we did $i)$ not run data sets that have multiple versions,
$ii)$ exclude data sets with less then 3 examples in each class $iii)$ remove data sets with
more than $10000$ instances or time series longer than $750$ measurements. As some of the classifiers  only work
with multi-class targets via \textit{1-vs-all} classification, we $iv)$ additionally excluded data sets with
more then 40 classes. In essence, we benchmark small and medium sized data sets with a moderate amount of different classes.

We add $7$ new algorithms and $6$ feature extraction methods which can be combined with arbitrary machine learning methods for scalar features (c.f. Table \ref{tab:benchmark}). Additionally we test $5$ classical machine learning methods, in order to obtain a broader perspective on expected performance if the functional nature of the data is ignored. As we benchmark default settings as well as tuned algorithms, in total $80$ different algorithms are evaluated across all data sets.
When combining feature extraction and machine learning methods, we fuse the learning algorithm and the preprocessing, thus treating them as a pipeline where data is internally transformed before applying the learner. This allows us to jointly tune the hyperparameters of learning algorithm and preprocessing method. The respective defaults and parameter ranges can be obtained from Table \ref{tab:parsets_fda_extractors} (feature extractors) and Table \ref{tab:ps_algo_tuned} (learning algorithms). More detailed description of the hyperparameters can be obtained from the respective packages documentation.

In order to generate train/test splits, and thus obtain an unbiased estimate of the algorithm's performance, we use stratified sub-sampling. We use 20 different train/test splits for each data set in order to reduce variance and report the average. For tuned models, we use use nested cross-validation \citep{bischl2012} to ensure unbiased estimates, where the outer loop is again subsampling with 20 splits, and the inner resampling for tuning is a 3-fold (stratified) cross-validation. All compared 80 algorithms are presented exactly the same index sets for the 20 train-test outer subsampling splits. 

Mean misclassification error (MMCE) is chosen as a measure of predictive performance in order to stay consistent with \citet{bagnall2017great}. Other measures, such as area under the curve (AUC) require predicted probabilities and do not trivially extend to multi-class settings.

While \citet{bagnall2017great} tune all algorithms across a carefully handcrafted grid, we use  \textit{Bayesian optimization} \citep{snoek12}. In order to stay comparable, we analogously fix the amount of tuning iterations to 100.

We use \pkg{mlrMBO} \citep{mlrMBO} in order to perform Bayesian optimization of the hyperparameters of the respective algorithm.
Additionally, in order to scale the method to a larger amount of data sets and machines, the R-package \pkg{batchtools} (\citet{batchtools1}, \citet{batchtools2}) is used. This enables running benchmark experiments on high-performance clusters. For the benchmark experiment, a job is defined as re-sampling of a single algorithm (or tuning thereof) on a single version of a data set. This allows for parallelization to an arbitrary number of CPU's, while at the same time guaranteeing reproducibility. The code for the benchmark is available from \url{https://github.com/compstat-lmu/2019\_fda_benchmark} for reproducibility.

\subsection{Results}

%
%
%
%

This Section tries to answer the questions posed in section \ref{sec:intro}. We evaluate $i)$ various machine learning algorithms in combination with feature extraction, $ii)$ classical time series classification approaches, $iii)$ the effect of tuning hyperparameters for several methods, and $iv)$ try to give recommendations with respect to which algorithm(s) to choose for new classification problems.

Algorithms evaluated in this benchmark have been divided into three groups: 
Algorithms specifically tailored to functional data, \textit{classical} machine learning algorithms without feature extraction and \textit{classical} machine learning algorithms in combination with feature extraction. 

\subsubsection{Algorithms for functional data}
\begin{figure}[h]
\centering
    \includegraphics[width=0.9\textwidth]{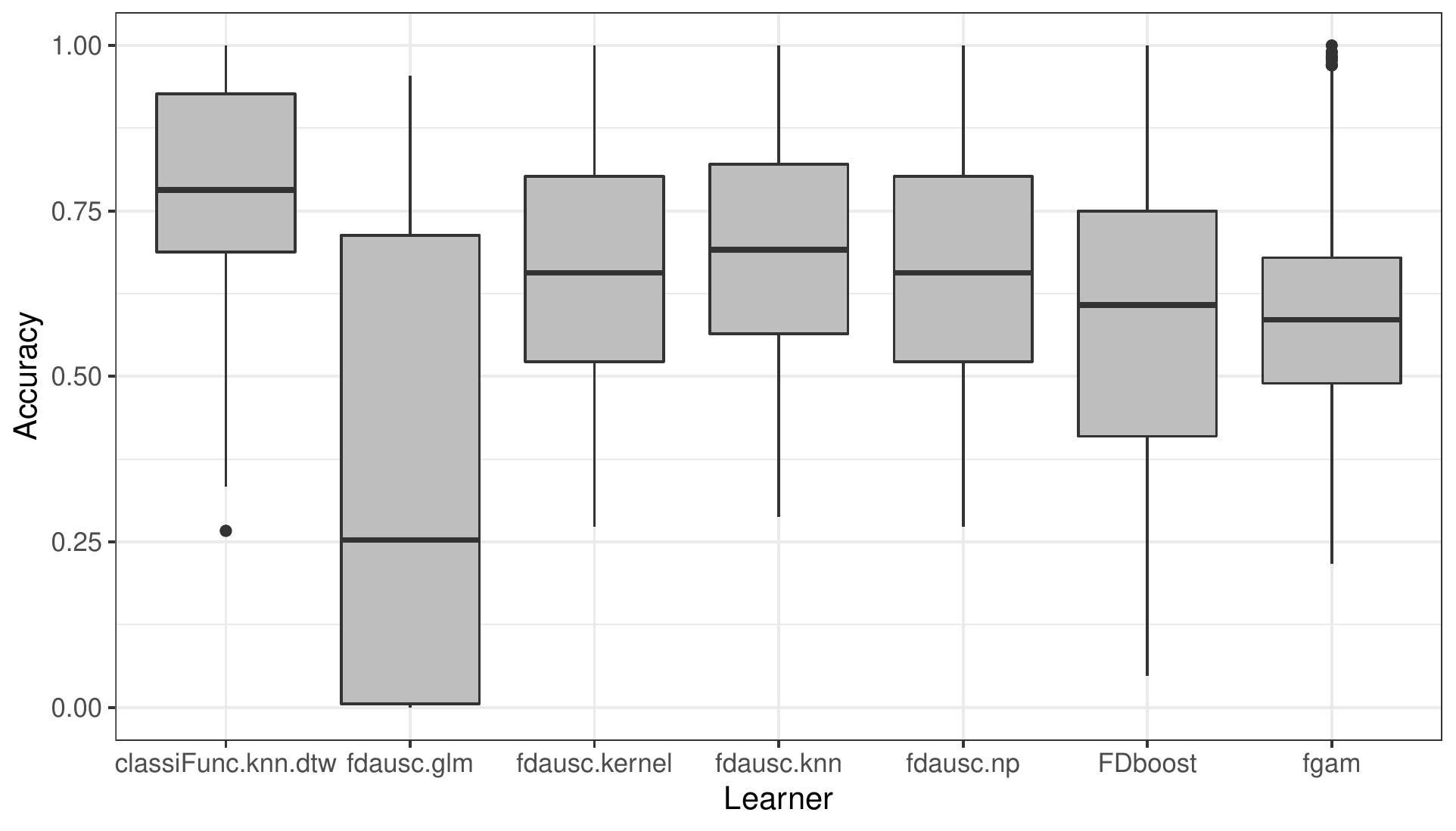}
    \caption{Performances for functional data analysis algorithms in default settings (untuned) across all $51$ data sets.}
\centering
\label{fig:fda_algos}
\end{figure}

Performances of algorithms specifically tailored to functional data analysis can be obtained from Figure \ref{fig:fda_algos}. The \textit{k-nearest neighbors} algorithm from package \pkg{classiFunc} \citep{classiFunc} in combination with dynamic time warping \citep{fastdtw} distance seems to perform best across data sets. It is also considered a \textit{strong baseline} in \citet{bagnall2017great}.

\subsubsection{Machine Learning algorithms with feature extraction}

\begin{figure}[ht!]
\centering
    \includegraphics[width=0.95\textwidth]{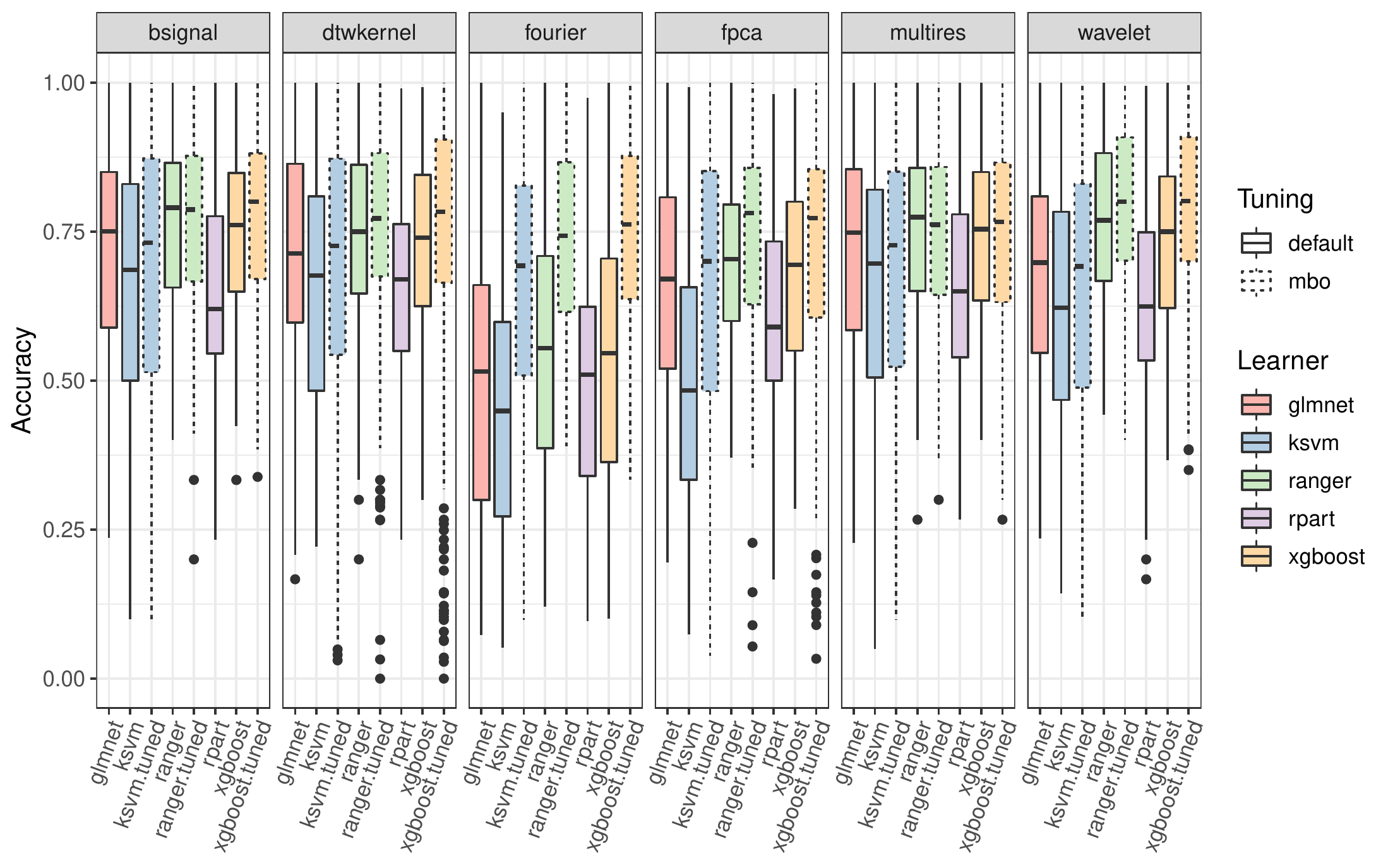}
    \caption{Results for feature extraction-based machine learning algorithms with default and tuned (MBO) hyperparameters across $51$ data sets. Hyperparameters are tuned jointly for learner and feature extraction method.}
\centering
\label{fig:ft_tuned_default}
\end{figure}

Performances of different machine learning algorithms in combination with feature extraction with and without tuning can be obtained from Figure \ref{fig:ft_tuned_default}.

We conclude that feature extraction using splines (bsignal) and wavelets as well as extracting dynamic time warping distances works well when combined with conventional machine learning algorithms, even at their default hyper-parameters. Among the learners, random forests, especially in combination with \textit{bsignal} show quite advantageous performance. In addition, we find an obvious improvement from hyper-parameter tuning for the Fourier feature extraction. In terms of learners, random forest and gradient boosted tree learners (\texttt{xgboost}) perform better than support vector machines.  
 
\subsubsection{Machine Learning algorithms without feature extraction}

\begin{figure}[h]
\centering
    \includegraphics[width=0.95\textwidth]{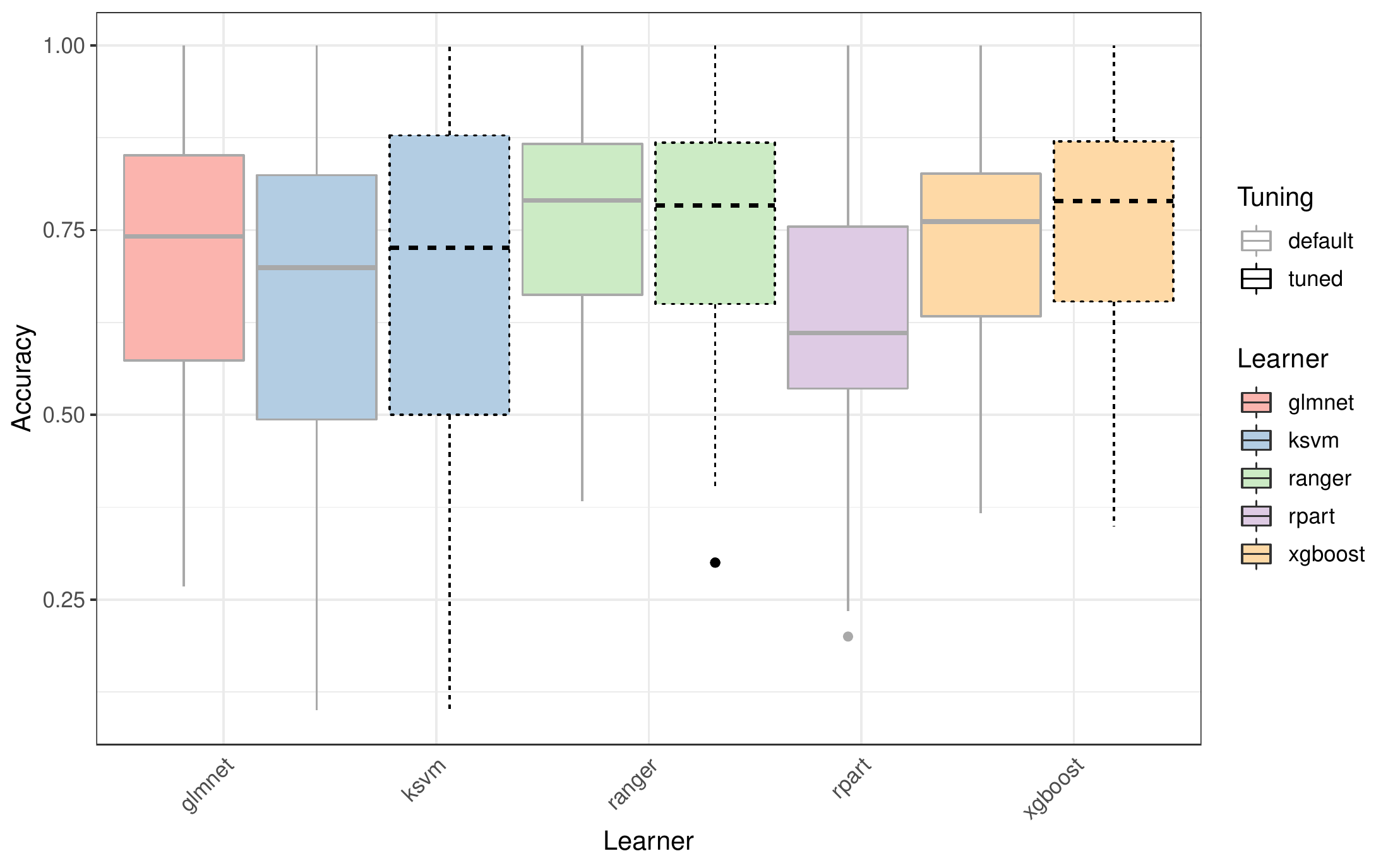}
    \caption{Performance of non-functional machine learning algorithms across $51$ data sets applied directly to functional data with and without tuning.}
\centering
\label{fig:tune}
\end{figure}

Additionally, we aim to provide some insight with regards to the performance of machine learning algorithms that ignore the functional nature of our data.  
Figure \ref{fig:tune} provides an overview over the performance of different machine learning algorithms that are often used together with traditional tabular data. Performances in this figure are obtained from algorithms directly applied to the functional data without any additional feature extraction.
The widely used gradient boosting (\texttt{xgboost}) and random forest (\texttt{ranger}) implementations seem to work reasonably well for functional data even without additional feature extraction.

\subsubsection{The effect of tuning hyperparameters}
From our experiments, we conclude, that tuning hyperparameters of machine learning algorithms in general has a non-negligible effect on the performance. Using Bayesian optimization in order to tune algorithm hyperparameters on average yielded an absolute increase in accuracy of  5.4\% across data sets and learners.

\begin{figure}[h]
\centering
    \includegraphics[width=0.9\textwidth]{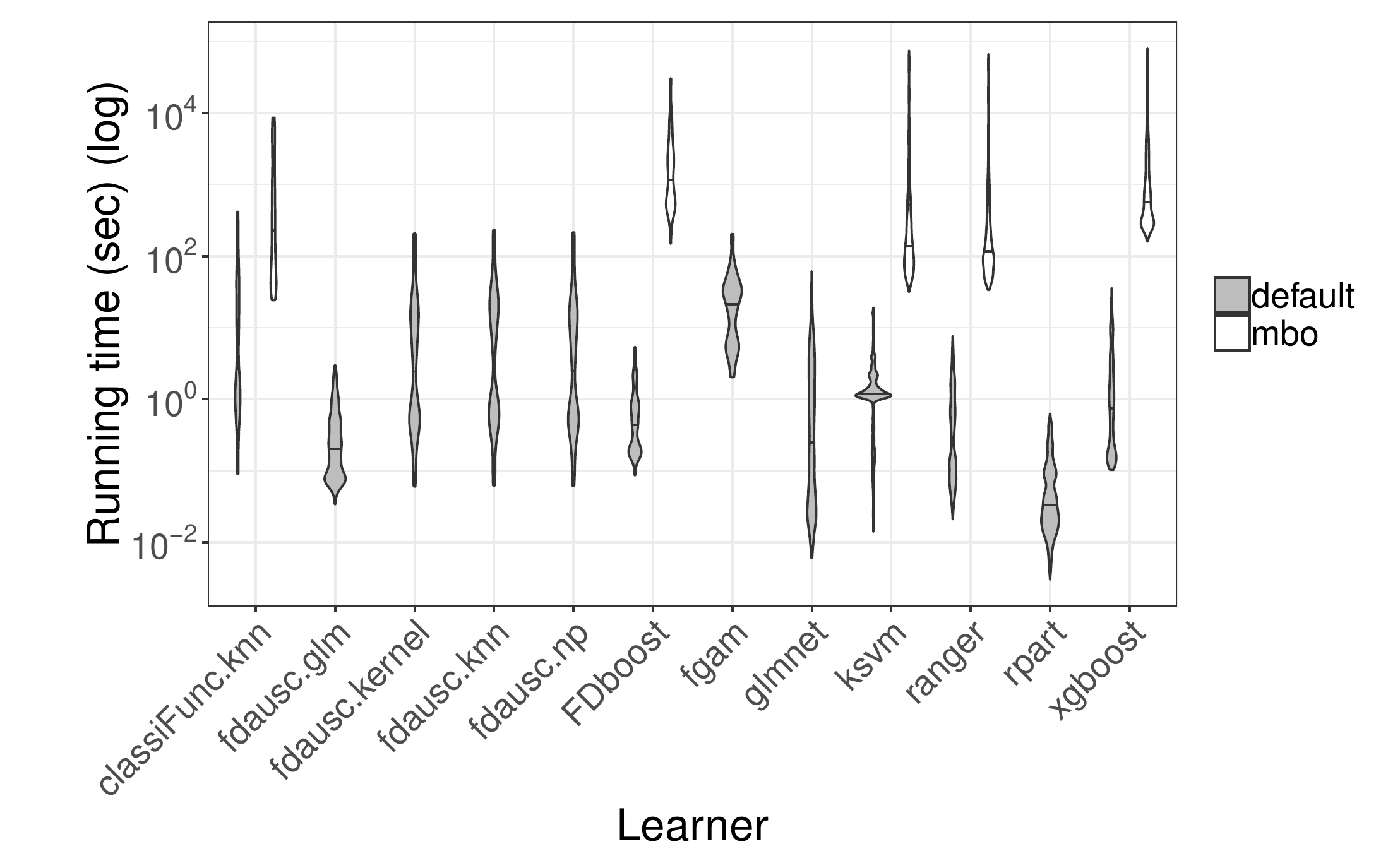}
    \caption{Comparison of running time for the different learner classes with default and tuned hyperparameters across $51$ data sets. A log transformation on the running time in seconds is applied, and the mean running time is visualized for each stratification as a horizontal line within the violin plot.}
\centering
\label{fig:rntime}
\end{figure}

Figure~\ref{fig:rntime} displays the aggregated time over all data sets, taking into account the time required for hyperparameter tuning. 
All experiments have been run on equivalent hardware on high-performance computing infrastructure. Due to fluctuations in server load, this does not allow for an exact comparison with respect to computation time, but we hope to achieve comparable results as we repeatedly evaluate on sub-samples. Note that we restrict the \textit{tuning} to $3$ algorithms where tuning traditionally leads to higher performances.\footnotemark\footnotetext{Additionally, we find significantly improved performance for tuned FDboost in Figure \ref{fig:fda_algos}}

\subsubsection{Top 10 Algorithms and recommendations}

Table \ref{tab:top10} showcases the top $10$ algorithms from the benchmark in terms of average rank in predictive accuracy across data sets. With this list, we aim to provide some initial understanding of the performance of different algorithms and feature extraction methods. Note that this list by no means reflects performance on future data sets, but might serve as an indicator, of which algorithms one might want to try first given computational constraints.

\begin{table}
\caption{Top 10 algorithms by average rank across all data sets. Percent Accuracy describes the fraction of the maximal accuracy reached for each task.}
\label{tab:top10}
\centering
\begin{tabular}{|l|r|r|}
  \hline
Algorithm Setting & Accuracy \% & Average Rank \\ 
  \hline
ranger\_wavelet\_tuned & 0.92 & 12.90 \\ 
  xgboost\_wavelet\_tuned & 0.92 & 14.45 \\ 
  ranger\_bsignal\_tuned & 0.90 & 15.02 \\ 
  knn\_dtw\_tuned & 0.92 & 15.22 \\ 
  ranger\_none\_default & 0.90 & 15.59 \\ 
  ranger\_bsignal\_default & 0.89 & 15.71 \\ 
  ranger\_wavelet\_default & 0.90 & 16.33 \\ 
  knn\_dtw\_default & 0.92 & 16.43 \\ 
  xgboost\_bsignal\_tuned & 0.90 & 17.57 \\ 
  ranger\_none\_tuned & 0.89 & 18.49 \\ 
   \hline
\end{tabular}
\end{table}

We observe that wavelet extraction in combination with either ranger or xgboost seems to be very strong. They obtain an average rank of 12.90 and 14.45 (out of 80) respectively. Dynamic time warping distances for k-nearest neighbors indeed seems to be a strong baseline, even without tuning. Another strong feature extraction method seems to be the extraction of B-spline features.
Using the $10$ algorithms above allows us to obtain an accuracy within $5\%$ of the maximum on $49$ of the $51$ data sets.

If the only criterion for model selection is predictive performance, (tuned) machine learning models in combination with feature extraction is a competitive baseline. This class of methods achieves within 95\% of the optimal performance on $47$ out of $51$ data sets, while they include the best performing classifier in $35$ cases.

\begin{table}[ht]
\centering
\begin{tabular}{lllll}
  \toprule
id & type & values & def. & trafo \\ 
  \midrule
  \textbf{bsignal} & & & & \\
  bsignal.knots & int & \{3,...,500\} & 10 & - \\ 
  bsignal.df & int & \{1,...,10\} & 3 & - \\ 
  \textbf{multires} & & & & \\
  res.level & int & \{2,...,5\} & - & - \\ 
  shift & num & [0.01,1] & - & - \\
  \textbf{pca} & & & & \\
  rank. & int & \{1,...,30\} & - & - \\ 
  \textbf{wavelets} & & & & \\
  filter & chr & d4,d8,d20,la8,la20,bl14,bl20,c6,c24 & - & - \\ 
  boundary & chr & periodic,reflection & - & - \\ 
  \textbf{fourier} & & & & \\
  trafo.coeff & chr & phase,amplitude & - & - \\ 
  \textbf{dtwkernel} & & & & \\
  ref.method & chr & random,all & random & - \\ 
  n.refs & num & [0,1] & - & - \\ 
  dtwwindow & num & [0,1] & - & - \\ 
   \bottomrule
\end{tabular}
\caption{Parameter spaces and default settings for feature extraction methods.}
\label{tab:parsets_fda_extractors}
\end{table}

\subsubsection{Comparison to classical time series classification}

\begin{figure}[ht]
\centering
    \includegraphics[width=0.6\textwidth]{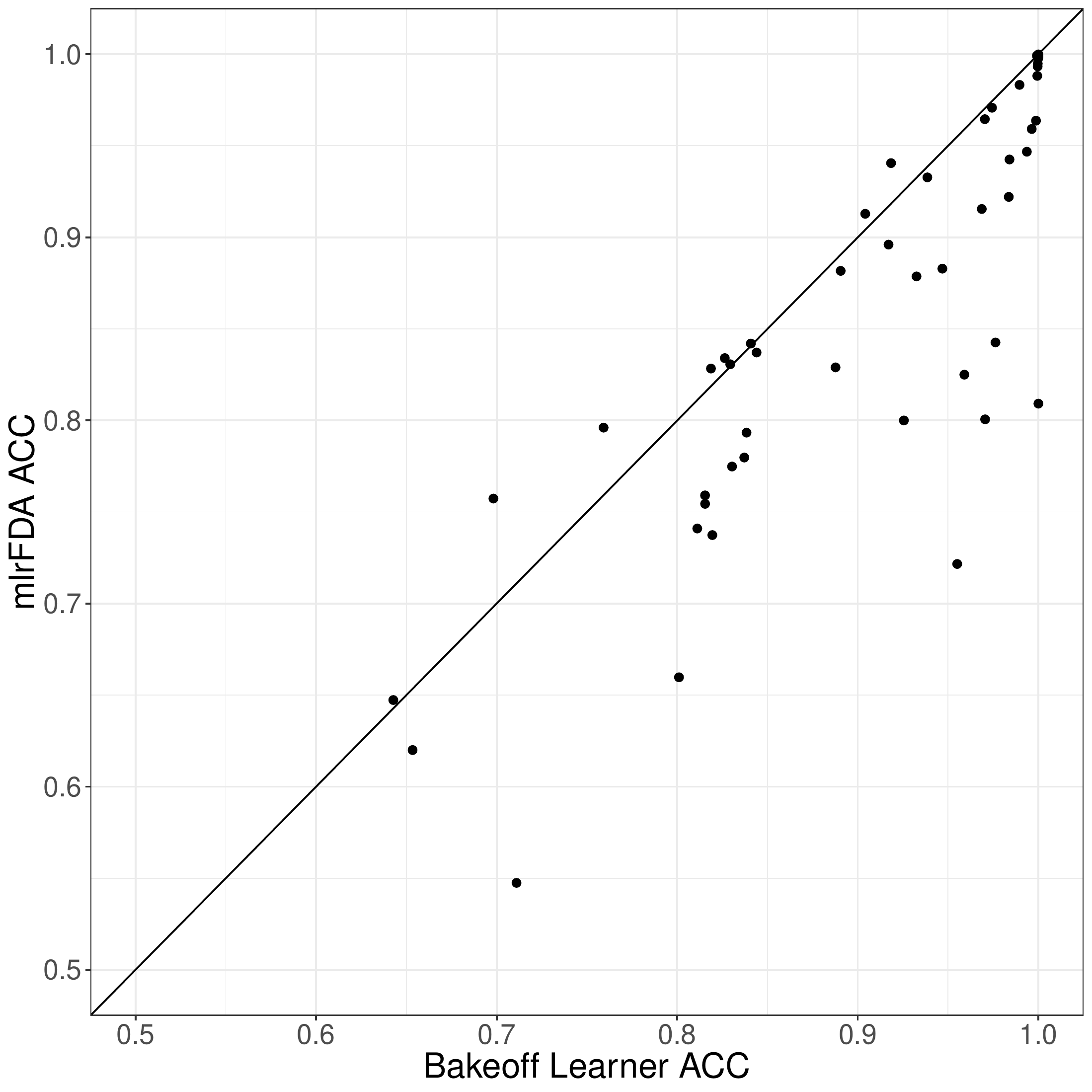}
    \caption{Comparing accuracy between our \pkg{mlrFDA} learners and the classical time series classification algorithms in \cite{bagnall2017great}. For each data set, only the best accuracy for each of the two benchmarks is shown. We observe that for 9 of the evaluated data sets the classification performance can directly be improved solely by applying our \pkg{mlrFDA} learners, while we perform on par with the classical time series classification algorithms (when rounding to 3 decimal digits) on two data sets. }
\centering
\label{fig:bakeoffvsmlr}
\end{figure}

Even though the main purpose of this paper is not a direct comparison with the results from \cite{bagnall2017great}, we can use our results to show that applying functional data approaches and classical machine learning approaches together with feature extraction can still improve classification accuracy compared to current state-of-the-art time series classification methods. \\
In the experiments we conducted, the methods described in this paper improved accuracy on 9 out of the 51 data sets which is displayed in Figure~\ref{fig:bakeoffvsmlr}. The 9 data sets and the corresponding best learner are displayed in Table~\ref{tab:bestmlrFDA}. For each data set, only the best reached accuracy for both sets of algorithms is displayed. 

\begin{table}[ht]
\centering
\begin{tabular}{llr}
  \hline
 Name & Algorithm\_Setting & Accuracy \\ 
  \hline
 Beef & xgboost\_wavelet\_tuned & 0.83 \\ 
   ChlorineConcentration & ksvm\_none\_tuned & 0.91 \\ 
   DistalPhalanxOutlineAgeGroup & ranger\_none\_default & 0.83 \\ 
  DistalPhalanxOutlineCorrect & ranger\_dtwkernel\_default & 0.83 \\ 
  DistalPhalanxTW & ranger\_bsignal\_default & 0.76 \\ 
  Earthquakes & FDboost\_none\_default & 0.80 \\ 
  Ham & xgboost\_wavelet\_tuned & 0.84 \\ 
  InsectWingbeatSound & ranger\_wavelet\_default & 0.65 \\ 
  SonyAIBORobotSurface1 & ksvm\_wavelet\_default & 0.94 \\ 
   \hline
\end{tabular}
\caption{Data sets together with corresponding \pkg{mlrFDA} learner and accuracy for which our learners were able to improve accuracy in the conducted experiments. }
\label{tab:bestmlrFDA}
\end{table}

\begin{figure}[ht]
\centering
    \includegraphics[width=0.95\textwidth]{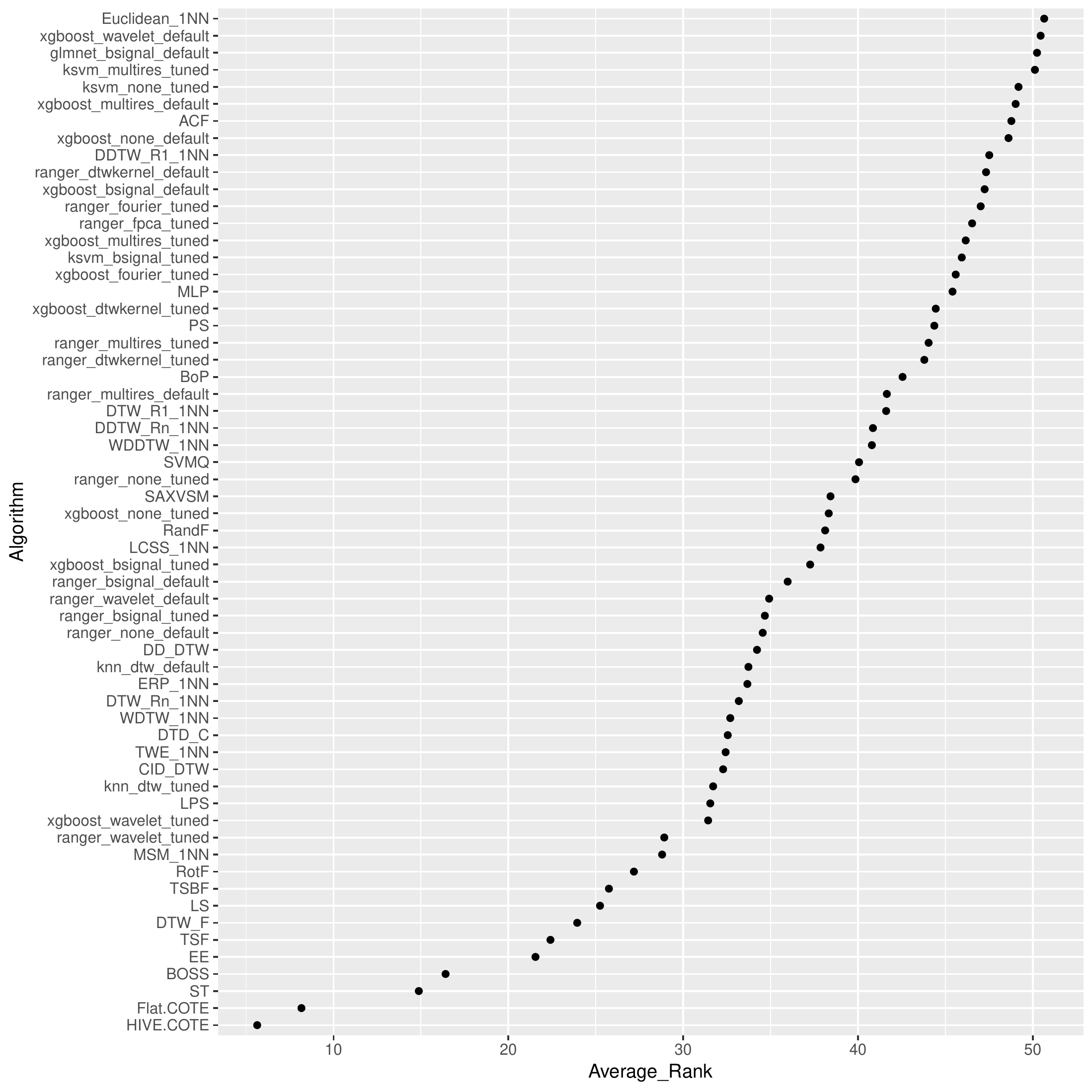}
    \caption{Comparing sorted average performance ranks between our \pkg{mlrFDA} learners (algorithm names in lower case) and the classical time series classification algorithms (algorithm names in capital) in \cite{bagnall2017great},  The mean rank of each individual learner over all $49$ data sets is displayed. Only the first half of all algorithms being compared are displayed here. We observe that the Ensemble Methods like HIVE.COTE, FLAT.COTE, ST, BOSS, EE occupy the top tier, while the rest of the rank space are interleaved by our \pkg{mlrFDA} algorithms and algorithms from \cite{bagnall2017great}
    }
\centering
\label{fig:ranking}
\end{figure}

Additionally, we evaluate how our learners rank in comparison to the individual bake-off algorithms from \cite{bagnall2017great}. 
The algorithm which performs best on a data set obtains the rank 1. The mean rank of the individual learners over all $49$ data sets (we take the intersection of the data sets from our benchmark and the ones from \cite{bagnall2017great}).
The average sorted ranks for the top 50\% algorithms are displayed in Figure ~\ref{fig:ranking}. 
We observe that the ensemble methods get the top ranks, which is no surprise, as for instance the COTE algorithm \cite{bagnall2015time} internally combines several classifiers from 4 different time series domains.

However, compared to the classical time series algorithms from  \cite{bagnall2017great} with the ensemble methods removed, our functional data algorithms obtain an overall good rank in accuracy performance, interleaved with the algorithms from \cite{bagnall2017great}. Note that the benchmarks are not exactly comparable due to minor differences in the benchmark setup, and 
we instead only include their reported results. 


\begin{table}[]
    \centering
\begin{tabular}{lllll}
  \textbf{parameter} & \textbf{type} & \textbf{values} & \textbf{default} & \textbf{trafo} \\
  \toprule
\textbf{ksvm} & & & & \\
  \toprule
    C & num & [-15,15] & - & 2\verb|^|x \\ 
    sigma & num & [-15,10] & - & 2\verb|^|x \\
     & & & & \\
   \bottomrule
\textbf{ranger} & & & & \\
  \toprule
    mtry.power & num & [0,1] & - & $p^x$ \\ 
  min.node.size & num & [0,0.99] & - & $2\verb|^|(log2(n) * x)$ \\ 
  sample.fraction & num & [0.1,1] & - & - \\ 
   & & & & \\
   \bottomrule
\textbf{xgboost} & & & & \\
  \toprule
nrounds & int & \{1,...,5000\} & 100 & - \\ 
  eta & num & [-10,0] & - & 2\verb|^|x \\ 
  subsample & num & [0.1,1] & - & - \\ 
  booster & chr & gbtree,gblinear & - & - \\ 
  max\_depth & int & \{1,...,15\} & - & - \\ 
  min\_child\_weight & num & [0,7] & - & 2\verb|^|x \\ 
  colsample\_bytree & num & [0,1] & - & - \\ 
  colsample\_bylevel & num & [0,1] & - & - \\ 
  lambda & num & [-10,10] & - & 2\verb|^|x \\ 
  alpha & num & [-10,10] & - & 2\verb|^|x \\ 
    & & & & \\
  \bottomrule
  \textbf{FDboost} & & & & \\
  \toprule
  mstop & int & \{1,...,5000\} & 100 & - \\ 
  nu & num & [0,1] & 0.01 & - \\ 
  df & num & [1,5] & 4 & - \\ 
  knots & int & \{5,...,100\} & 10 & - \\ 
  degree & int & \{1,...,4\} & 3 & - \\ 
   \bottomrule

\end{tabular}
    \caption{Parameter spaces and defaults used for tuning machine learning and functional data algorithms. In case no default is provided, package defaults are used.
    Additional information can be found in the respective packages documentation.}
    \label{tab:ps_algo_tuned}
\end{table}

\section{Summary and Outlook}

In this work, we provide a benchmark along with a software implementation that integrates the functionality of a diverse set of R-packages into a single user interface and API. Both contributions come with a multiplicity of benefits:

\begin{itemize} 
\item The user is not required to learn and deal with the vast complexity of the different interfaces the underlying packages expose. 
\item All of the existing functionality (e.g., preprocessing, resampling, performance measures, tuning, parallelization) of the \pkg{mlr} ecosystem can now be used in conjunction with already existing algorithms for functional data.
\item We expose functionality that allows us to work with functional data using \textit{traditional} machine learning methods via feature extraction methods.
\item Integration of additional preprocessing methods or models is (fairly) trivial and automatically benefits from the full \pkg{mlr} ecosystem.
\end{itemize}

In order to obtain a broader overview of the performance of the integrated methods, we perform a large benchmark study. This allows users to get an initial overview of potential performances of the different algorithms. Specifically, 
\begin{itemize}
    \item We open up new perspectives for time series classification tasks by incorporating methods from functional data analysis, as well as feature transformations combined with conventional machine learning models.
    \item Based on the large scale benchmark, we conclude that many learners have competitive performance (Figure \ref{fig:bakeoffvsmlr})  and additionally offer the interpretability of many functional data analysis methods. Our toolbox serves as a strong complement and alternative to other time series classification software.
    \item 
    The presented benchmark study uses state-of-the-art Bayesian optimization for hyperparameter optimization, which results in significant improvements over models that are not tuned. This kind of hyperparameter tuning is easy to do with \pkg{mlrFDA}. Tuning, albeit heavily influencing performance is often not investigated. Our benchmark closes this gap in existing literature.

    \item We find that extracting vector valued features and feeding them to a conventional machine learning model can often form competitive learners.
    
    \item The pareto-optimal set in terms of performance on each data set contains $23$ different algorithm$-$feature-extraction combinations. Our toolbox $i)$ offers the same API for all methods and $ii)$ allows to automatically search over this space, and thus allows users to obtain optimal models without knowing all underlying methods. 
\end{itemize}
Concerning the questions we proposed at the beginning of the paper, we draw the following conclusions:

\begin{itemize}
    \item Tuning only a subset of the presented learners and feature extractions, i.e., the methods listed in Table \ref{tab:top10},
    is sufficient to achieve good performances on almost all data sets in our benchmark.
    
     \item A simple random forest without any preprocessing can also be a reasonable baseline for time series data. It achieves an average rank of $15.59$ (top 4) in our benchmark.
    
    \item Most algorithms for functional data (e.g., \texttt{FDboost}) do not perform well in our benchmark study. As those algorithms are fully interpretable and offer statistically valid coefficients, they can still be useful in some applications, and should thus not be
    ruled out.
    
    \item Feature extraction techniques, such as b-spline representations (\textit{bsignal}) and wavelet extraction work well in conjunction with machine learning techniques for vector valued features such as \texttt{xgboost} and \texttt{random forest}.
    
    \item Tuning leads to an average reduction in absolute MMCE of $3.59\%$ (ranger), $5.69\%$ (xgboost), $7.78\%$ (ksvm) (across feature extraction techniques) and $11\%$ (FDboost). This holds for all feature extraction techniques, where improvements range from $1.12\%$ \textit{multires} to $20.3\%$ \textit{fourier}.
\end{itemize}

In future work we will continue to expand the available toolbox along with a benchmark of new methods, and provide the R community a wider range of methods that can be used for the analysis of functional data. This includes not only integrating many already available packages, and as a result to enable preprocessing operations such as smoothing (e.g., \pkg{fda} \citep{fda}) and alignment (e.g., \pkg{fdasrvf} \citep{fdasrvf} or \pkg{tidyfun} \citep{tidyfun}), but also to explore and integrate advanced imputation methods for functional data.
Further work will also extend the current implementation to support data that is measured on unequal or irregular grids.
Additionally, we aim to implement some of the current state-of-the art machine learning models from the time series classification bake-off \citep{bagnall2017great}, such as the \textit{Collective of Transformation-Based Ensembles} (COTE) \citep{COTE}. This enables researchers to use and compare with current state-of-the-art methods.

\subsection*{Acknowledgements}

This work has been funded by the German Federal Ministry of Education and Research (BMBF) under Grant No. 01IS18036A. The authors of this work take full responsibilities for its content.

\newpage

\bibliographystyle{elsarticle-num-names}
\bibliography{paper_fda}

\newpage

\appendix

\section{API overview}

For the interested reader, we introduce a brief overview of the API
and functionality.

\subsection{Representing functional data in mlrFDA}
\label{sec:fda_models}

\begin{figure}[!htbp]
\centering
    \includegraphics[width=0.6\textwidth]{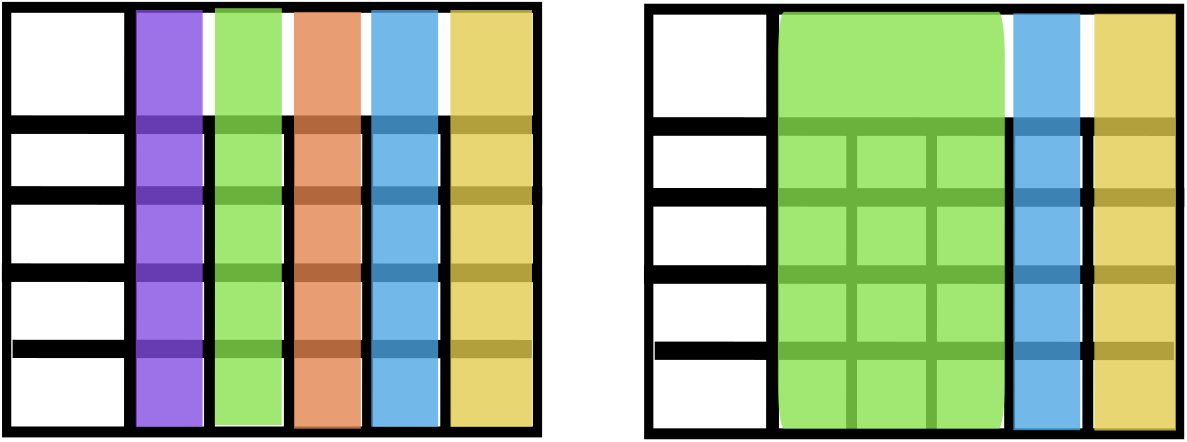}
\caption{Schematic comparison of non-functional and functional data representation in mlrFDA. The green feature is a functional feature spanning multiple columns.}
\label{fig:database}
\end{figure}

A sketch of the data structure we use to represent functional data can be found in the right part of Figure \ref{fig:database}. We assume a data set consists of data for $N$ observational units, organized in rows of features, where one row contains all observed features for one observational unit, i.e., each row typically contains several functional and/or scalar covariates. For a classical, non-functional data set, the $P$ features are single columns (as depicted in the left part of Figure \ref{fig:database}). A functional data set, on the other hand, consists of single-column scalar features as well as functional features of different length for each functional covariate, each represented by multiple adjacent and connected columns. Each of these columns contains the evaluations of the functional feature at a certain argument value for all observational units (right part of Figure \ref{fig:database}). 

As an example which will be used throughout the remainder of this paper, we use the \code{fuelSubset} data set from  package \pkg{FDboost}, see also Figure \ref{fig:fdboost}. It contains a numeric target variable \code{heatan}, the fuel's heating value, a scalar feature \code{h20}, the fuel's water content, and two functional features \code{NIR} and \code{UVVIS}, measured at 231 and 129 wavelengths, respectively. To start with a clean sheet, we create a \code{data.frame} containing all features as separate columns.
 
\begin{CodeChunk}
\begin{CodeInput}
R> library(mlr)
R> library(FDboost)
R> df = data.frame(fuelSubset[c("heatan", "h2o", "UVVIS", "NIR")])
\end{CodeInput}
\end{CodeChunk}

The first step when setting up an experiment in any analysis is to make the data accessible for the specific algorithms that will be applied. In \pkg{mlr}, the data itself, and additional information, such as which column corresponds to the target variable is stored as a \code{Task}, requiring the input data to be of type \code{data.frame}. \\

The list of column positions of the functional features is then passed as argument \code{fd.features} to \code{makeFunctionalData()}, which returns an object of type \code{data.frame} in which the columns corresponding to each  functional feature are combined into \code{matrix} columns.
\footnote{As an alternative, a \textit{list} of the column names containing the functional features is also valid as argument to \textit{fd.features}, which is especially useful if columns are already labeled.} \\

\begin{CodeChunk}
\begin{CodeInput}
R> fd.features = list("UVVIS" = 3:136, "NIR" = 137:367)
R> fdf = makeFunctionalData(df, fd.features = fd.features)
R> str(fdf)
'data.frame':	129 obs. of  4 variables:
 \$ heatan: num  26.8 27.5 23.8 18.2 17.5 ...
 \$ h2o   : num  2.3 3 2 1.85 2.39 ...
 \$ UVVIS : num [1:129, 1:134] 0.145 -1.584 -0.814 -1.311 -1.373 ...
  ..- attr(*, "dimnames")=List of 2
  .. ..\$ : NULL
  .. ..\$ : chr  "UVVIS.1" "UVVIS.2" "UVVIS.3" "UVVIS.4" ...
 \$ NIR   : num [1:129, 1:231] 0.2818 0.2916 -0.0042 -0.034 -0.1804 ...
  ..- attr(*, "dimnames")=List of 2
  .. ..\$ : NULL
  .. ..\$ : chr  "NIR.1" "NIR.2" "NIR.3" "NIR.4" ...
\end{CodeInput}
\end{CodeChunk}

We additionally specify the name \code{"fuelsubset"} and the target variable \code{"heatan"}.
The structure of the functional \code{Task} object is rather similar to the non-functional \pkg{Task}, with the additional information \code{functionals}, which states how many functional features are present in the underlying data.

\begin{CodeChunk}
\begin{CodeInput}
R> tsk1 = makeRegrTask("fuelsubset", data = fdf, target = "heatan")
R> print(tsk1)
	Supervised task: fuelsubset
	Type: regr
	Target: heatan
	Observations: 129
	Features:
	   numerics     factors     ordered functionals 
	          1           0           0           2 
	Missings: FALSE
	Has weights: FALSE
	Has blocking: FALSE
	Has coordinates: FALSE
\end{CodeInput}
\end{CodeChunk}

After defining the task, a \code{learner} is created by calling \code{makeLearner}. This contains the algorithm that will be fitted on the data in order to obtain a model.
Currently, \pkg{mlrFDA} supports both functional regression and functional classification.  A list of supported learners can be found in Table \ref{tab:benchmark}
\subsection{Machine Learning and Feature Extraction}
\label{sec:fda_featextract}

Classical machine learning algorithms do not take into account the characteristics of functional data and treat the input data as vector valued features. Without additional preprocessing, this typically yields poor performance on, as the models cannot exploit the lower intrinsic dimensionality of the functional covariates nor the fact that they represent observations over a continuum.

In \pkg{mlrFDA}, classical algorithms can be applied to functional data, however, a warning message will be displayed. In our example, we train a partitioning tree on the functional data, while ignoring the functional structure.

\begin{CodeChunk}
\begin{CodeInput}
R> rpart.lrn = makeLearner("regr.rpart")
R> m = train(learner = rpart.lrn, task = tsk1)
   Functional features have been converted to numerics
\end{CodeInput}
\end{CodeChunk}

For conventional learning algorithms to work well on functional data, informative scalar features need to be extracted from the functional features.

Feature extraction is applied in practice for a multiplicity of reasons, as it often not only reduces the dimensionality of the resulting problem, but also allows researchers to make use of domain knowledge, for example by hand-crafting features from measurements of continuous processes. Examples for this include deriving features that allow for sleepiness detection \citep{driver_sleepiness}, or by extracting features from electro-cardiogram data in order to detect emotions \citep{eeg_emotions}. The resulting features often have a much lower dimensionality, which often improves fitted models. 
Other preprocessing methods for functional or time series data include extracting general purpose features such as wavelet coefficients \citep{wavelets, wavelet_timeseries}, principal component scores or Fourier coefficients. The resulting scalar features can then be used with different machine learning methods such as $k$-nearest neighbors.

In the following section, we showcase the feature extraction procedure using general purpose features as an example. We want to emphasize that it is also easily possible to write custom feature extraction methods using the \code{makeFeatureExtractionMethod} function.

\begin{table}
    \centering
    \scalebox{0.8}{
    \begin{tabular}{|c|c|c|}
\hline
Name & Function & Package   \\
\hline
Discrete Wavelet Transform & \code{extractFDAWavelets()} & \pkg{wavelets} \\
Fast Fourier Transform & \code{extractFDAFourier()} & \pkg{stats} \\

Principal Component Analysis & \code{extractFDAPCA()} & \pkg{stats} \\
B-Spline Features & \code{extractFDABsignal()} & \pkg{FDboost}\\
Multi-Resolution Feature Extraction&\code{extractFDAMultiResFeatures()} & \pkg{-} \\
Time Series Features & \code{extractFDATsfeatures()} & \pkg{tsfeatures} \\
Dynamic Time-Warping Kernel & \code{extractFDADTWKernel()} & \pkg{rucrdtw} \\
\hline
    \end{tabular}}
    \caption{Feature extraction methods currently implemented in \pkg{mlrFDA} and underlying packages}
    \label{tab:feature_extractors}
\end{table}

In our example, we extract the Fourier coefficients from the functional feature \code{UVVIS}, and principal component scores from the second functional feature \code{NIR}
in order to transform the original task with functional data into a conventional task.

\begin{CodeChunk}
\begin{CodeInput}
R> feat.methods = list("UVVIS" = extractFDAFourier(),
"NIR" = extractFDAPCA())
R> extracted = extractFDAFeatures(tsk, feat.methods = feat.methods)
R> extracted

  $task
  Supervised task: fuelsubset
  Type: regr
  Target: heatan
  Observations: 129
  Features:
     numerics     factors     ordered functionals 
          137           0           0           0 
  Missings: FALSE
  Has weights: FALSE
  Has blocking: FALSE
  Has coordinates: FALSE
  
  $desc
  Extraction of features from functional data:
  Target: heatan
  Functional Features: 2; Extracted features: 2
\end{CodeInput}
\end{CodeChunk}

As an alternative, the feature extraction can be applied in a wrapper method \\ \code{makeExtractFDAFeatsWrapper()}. In general, a wrapper combines a learner method with another method, thereby creating a new learner that can be handled like any other learner. In our case, a classical machine learning method is combined with the data preprocessing step of feature transformation from functional to non-functional data. 

\begin{CodeChunk}
\begin{CodeInput}
R> wrapped.lrn = makeExtractFDAFeatsWrapper("regr.rpart",
 feat.methods = feat.methods)
\end{CodeInput}
\end{CodeChunk}

This is suitable for honest cross-validation of data-adaptive feature extraction methods like principal components. We can now cross-validate the learner created above using \pkg{mlr}'s \texttt{resample} function with 10-fold cross-validation.

\begin{CodeChunk}
\begin{CodeInput}
R> res = resample(learner = wrapped.lrn, task = tsk1, 
resampling = cv10)
\end{CodeInput}
\end{CodeChunk}

In the same way, we can \texttt{train} and \texttt{predict} on data, or \texttt{benchmark} multiple learners across multiple data sets.  Additionally, we can apply a \texttt{tuneWrapper} to our learner in order to automatically tune hyperparameters of the learner and the preprocessing method during cross-validation.

\nocite{rucrdtw}
\nocite{fastdtw}

\section{Data sets used in the Benchmark}
Table \ref{tab:datasets} contains all data sets used in the benchmark along with additional data properties.

\begin{table}[h]
\begin{small}
\singlespacing
\centering
\begin{tabular}{lrrrrl}
  \toprule
Name & Obs. & Classes & Length & Type & Split \\
  \midrule
Adiac & 781 &  37 & 176 & IMAGE & 0.50 \\
  ArrowHead & 211 &   3 & 251 & IMAGE & 0.17 \\
  Beef &  60 &   5 & 470 & SPECTRO & 0.50 \\
  BeetleFly &  40 &   2 & 512 & IMAGE & 0.50 \\
  BirdChicken &  40 &   2 & 512 & IMAGE & 0.50 \\
  Car & 120 &   4 & 577 & SENSOR & 0.50 \\
  CBF & 930 &   3 & 128 & SIMULATED & 0.03 \\
  ChlorineConcentration & 4307 &   3 & 166 & SIMULATED & 0.11 \\
  Coffee &  56 &   2 & 286 & SPECTRO & 0.50 \\
  Computers & 500 &   2 & 720 & DEVICE & 0.50 \\
  CricketX & 780 &  12 & 300 & MOTION & 0.50 \\
  DistalPhalanxOutlineAgeGroup & 539 &   3 &  80 & IMAGE & 0.74 \\
  DistalPhalanxOutlineCorrect & 876 &   2 &  80 & IMAGE & 0.68 \\
  DistalPhalanxTW & 539 &   6 &  80 & IMAGE & 0.74 \\
  Earthquakes & 461 &   2 & 512 & SENSOR & 0.70 \\
  ECG200 & 200 &   2 &  96 & ECG & 0.50 \\
  ECGFiveDays & 884 &   2 & 136 & ECG & 0.03 \\
  ElectricDeviceOn & 1008 &   2 & 360 & DEVICE & 0.63 \\ 
  EpilepsyX & 275 &   4 & 208 & HAR & 0.61 \\ 
  FaceAll & 2250 &  14 & 131 & IMAGE & 0.25 \\ 
  FacesUCR & 2250 &  14 & 131 & IMAGE & 0.09 \\ 
  Fish & 350 &   7 & 463 & IMAGE & 0.50 \\ 
  GunPoint & 200 &   2 & 150 & MOTION & 0.25 \\ 
  Ham & 214 &   2 & 431 & SPECTRO & 0.51 \\ 
  Herring & 128 &   2 & 512 & IMAGE & 0.50 \\ 
  InsectWingbeatSound & 2200 &  11 & 256 & SENSOR & 0.10 \\ 
  ItalyPowerDemand & 1096 &   2 &  24 & SENSOR & 0.06 \\ 
  LargeKitchenAppliances & 750 &   3 & 720 & DEVICE & 0.50 \\ 
  Lightning2 & 121 &   2 & 637 & SENSOR & 0.50 \\ 
  Lightning7 & 143 &   7 & 319 & SENSOR & 0.49 \\ 
  Meat & 120 &   3 & 448 & SPECTRO & 0.50 \\ 
  MedicalImages & 1141 &  10 &  99 & IMAGE & 0.33 \\ 
  MoteStrain & 1272 &   2 &  84 & SENSOR & 0.02 \\ 
  OSULeaf & 442 &   6 & 427 & IMAGE & 0.45 \\ 
  Plane & 210 &   7 & 144 & SENSOR & 0.50 \\ 
  RefrigerationDevices & 750 &   3 & 720 & DEVICE & 0.50 \\ 
  ScreenType & 750 &   3 & 720 & DEVICE & 0.50 \\ 
  ShapeletSim & 200 &   2 & 500 & SIMULATED & 0.10 \\ 
  SmallKitchenAppliances & 750 &   3 & 720 & DEVICE & 0.50 \\ 
  SonyAIBORobotSurface1 & 621 &   2 &  70 & SENSOR & 0.03 \\ 
  Strawberry & 983 &   2 & 235 & SPECTRO & 0.62 \\ 
  SwedishLeaf & 1125 &  15 & 128 & IMAGE & 0.44 \\ 
  SyntheticControl & 600 &   6 &  60 & SIMULATED & 0.50 \\ 
  ToeSegmentation1 & 268 &   2 & 277 & MOTION & 0.15 \\ 
  Trace & 200 &   4 & 275 & SENSOR & 0.50 \\ 
  TwoLeadECG & 1162 &   2 &  82 & ECG & 0.02 \\ 
  TwoPatterns & 5000 &   4 & 128 & SIMULATED & 0.20 \\ 
  UWaveGestureLibraryX & 4478 &   8 & 315 & MOTION & 0.20 \\ 
  Wafer & 7164 &   2 & 152 & SENSOR & 0.14 \\ 
  Wine & 111 &   2 & 234 & SPECTRO & 0.51 \\ 
  Yoga & 3300 &   2 & 426 & IMAGE & 0.09 \\ 
   \bottomrule
   \bottomrule
\end{tabular}
\end{small}
\centering
\caption{Data sets from the UCI Archive used in the benchmark.}
\label{tab:datasets}
\doublespacing
\end{table}

\section{Failed and missing experiments}
Experiments for some algorithm / data set combinations failed due to 
implementation details or algorithm properties.
In order to increase transparency, failed algorithms are listed here, and if available reasons for failure are provided. 

At the time of the benchmark, the implementation in the \pkg{tsfeatures} package was not stable enough to be included in the benchmark.

\begin{itemize}
    \item \texttt{classif.fgam} \\
    \textbf{Data sets:} BeetleFly, BirdChicken, Coffee, Computers, DistalPhalanxOutlineCorrect, Earthquakes, ECG200, ECGFiveDays, ElectricDeviceOn, GunPoint, Ham, Herring, ItalyPowerDemand, Lightning2, MoteStrain, ShapeletSim, SonyAIBORobotSurface1, Strawberry, ToeSegmentation1, TwoLeadECG, Wafer, Wine, Yoga \\
    \textbf{Reason}: Too few instances in some classes, such that p > n.
    
    \item \texttt{classif.fdausc.kernel and .np} 
    \textbf{Data sets:} ElectricDeviceOn, ShapeletSim

    \item \texttt{classif.fdausc.knn} \\
    \textbf{Data sets:} DistalPhalanxTW, EpilepsyX
\end{itemize}

\end{document}